\documentclass[lettersize,journal]{IEEEtran}
\usepackage{amsmath,amssymb,amsfonts}
\usepackage{algorithm}
\usepackage{array}
\usepackage[caption=false,font=normalsize,labelfont=sf,textfont=sf]{subfig}
\usepackage{textcomp}
\usepackage{stfloats}
\usepackage{url}
\usepackage{verbatim}
\usepackage{graphicx}
\usepackage{cite}
\usepackage[utf8]{inputenc} 
\usepackage[T1]{fontenc}   
\usepackage{hyperref}       
\usepackage{booktabs}     
\usepackage{amsfonts}     
\usepackage{nicefrac}      
\usepackage{microtype}    
\usepackage{xcolor}        
\usepackage{algpseudocode}
\usepackage{wrapfig}
\usepackage{caption}
\usepackage[numbers]{natbib}
\usepackage{orcidlink}
\usepackage{multirow}
\begin{document}

\title{TAP-ViTs: Task-Adaptive Pruning for On-Device Deployment of Vision Transformers}

\author{
        Zhibo Wang\orcidlink{0000-0002-5804-3279},~\IEEEmembership{Senior Member,~IEEE,}
        Zuoyuan Zhang\orcidlink{0000-0002-8853-0078},
        Xiaoyi Pang\orcidlink{0000-0002-2763-2695},\\
        Qile Zhang\orcidlink{0009-0008-7991-6471},
        Xuanyi Hao\orcidlink{0009-0008-9742-9116},
        Shuguo Zhuo\orcidlink{0000-0001-5200-7130},
        Peng Sun\orcidlink{0000-0001-6221-8142}
        
\thanks{Zhibo Wang, Zuoyuan Zhang, Xiaoyi Pang, Qile Zhang, Xuanyi Hao, Shuguo Zhuo, are with the State Key Laboratory of Blockchain and Data Security and the College of Computer Science and Technology, Zhejiang University, Hangzhou 310027, China. (E-mail: {zhibowang, zuoyuan2024, qlzhng, xyhao, shuguozhuo}@zju.edu.cn; xypang@whu.edu.cn).}  

% \thanks{Xiaoyi Pang is with the Hong Kong University of Science and Technology, Hong Kong SAR, China. (E-mail: xypang@whu.edu.cn).}
        
\thanks{Peng Sun is with the College of Computer Science and Electronic Engineering, Hunan University, Changsha 410082, China. (E-mail: psun@hnu.edu.cn).}        
}

        % <-this % stops a space



\maketitle

\begin{abstract}

Vision Transformers (ViTs) have demonstrated strong performance across a wide range of vision tasks, yet their substantial computational and memory demands hinder efficient deployment on resource-constrained mobile and edge devices. Pruning has emerged as a promising direction for reducing ViT complexity. However, existing approaches either (i) produce a single pruned model shared across all devices, ignoring device heterogeneity, or (ii) rely on fine-tuning with device-local data, which is often infeasible due to limited on-device resources and strict privacy constraints. As a result, current methods fall short of enabling task-customized ViT pruning in privacy-preserving mobile computing settings.
This paper introduces TAP-ViTs, a novel task-adaptive pruning framework that generates device-specific pruned ViT models without requiring access to any raw local data. Specifically, to infer device-level task characteristics under privacy constraints, we propose a Gaussian Mixture Model (GMM)-based metric dataset construction mechanism. Each device fits a lightweight GMM to approximate its private data distribution and uploads only the GMM parameters. Using these parameters, the cloud selects distribution-consistent samples from public data to construct a task-representative metric dataset for each device. Based on this proxy dataset, we further develop a dual-granularity importance evaluation–based pruning strategy that jointly measures composite neuron importance and adaptive layer importance, enabling fine-grained, task-aware pruning tailored to each device’s computational budget.
Extensive experiments across multiple ViT backbones and datasets demonstrate that TAP-ViTs consistently outperforms state-of-the-art pruning methods under comparable compression ratios. Notably, TAP-ViTs maintains high accuracy even when retaining only 70\% of the model parameters, and in several cases surpasses the original unpruned models. These results highlight the effectiveness, task adaptivity, and deployability of our framework for privacy-aware on-device ViT compression.

\end{abstract}

\begin{IEEEkeywords}
Pruning, Vision Transformers.
\end{IEEEkeywords}

\section{Introduction} \label{sec:Introduction}

\IEEEPARstart{V}{ision} Transformers (ViTs) \cite{ViT, DeiT} have recently emerged as a powerful alternative to traditional convolutional neural networks (CNNs) for a wide range of vision tasks\cite{cordonnier2019relationship}, such as image classification \cite{he2022masked}, object detection \cite{carion2020end}, and video segmentation \cite{wang2021end}. 
ViTs leverage self-attention mechanisms to capture long-range dependencies and global context more effectively, thereby achieving superior performance in various vision tasks. 
Despite their impressive performance, ViTs typically require considerable computational and memory resources, which limits their applicability on mobile and edge devices with constrained resources.

To mitigate these limitations, pruning \cite{lpvit, gvit, ovit, dcvit} has emerged as a key model compression technique to reduce the computational and memory footprint of ViTs, enabling their efficient deployment on resource-constrained end devices while preserving performance.
Existing ViT pruning methods typically rely on a fixed, publicly available metric dataset to guide the pruning process \cite{mdvit, upvit}. Usually, they evaluate neuron importance on this metric dataset using a single importance criterion and prune model parameters with a predefined, uniform ratio across all layers.
With such a strategy, they generate uniform pruned ViTs for all end devices, failing to accommodate end devices'  local data distribution and personalized task requirements. A straightforward solution involves fine-tuning the pruned ViTs on local data to meet the requirements of specific downstream tasks. However, end devices usually have constrained computational resources and thus are unable to perform training. Meanwhile, they may be unwilling to share their local data with the cloud due to privacy concerns. In this context, it becomes crucial to customize pruned ViTs for end devices without accessing their original raw data. 

In this paper, we aim to design a pruning method to craft task-adaptive pruned ViTs for on-device deployment without requiring access to raw local data. 
However, achieving this goal presents two key challenges. 
First, without direct access to devices' local data, accurately measuring task requirements and adapting the pruning strategy accordingly becomes a critical bottleneck. 
How to construct approximate task-specific metric datasets to reflect the individual task requirements is a big challenge.
Second, given the high redundancy and architectural complexity of ViTs, there is a pressing need for an efficient pruning strategy that can accurately identify the neurons and layers most relevant to the target task. This demands not only accurate identification of critical components from thousands of redundant parameters but also a carefully balanced pruning strategy that maintains task-specific performance. It is challenging to achieve such precision under strict efficiency constraints.

To address these challenges, we propose a novel task-adaptive pruning framework tailored for ViTs, called TAP-ViTs, which enables task-adaptive and efficient deployment of ViTs on heterogeneous end devices without requiring access to private local data. Specifically, to perceive task characteristics without direct access to device-specific data, we introduce a GMM-based metric dataset construction method. Each device locally fits a Gaussian Mixture Model (GMM) to approximate the distribution of its private data. Instead of sharing raw local data, only the lightweight GMM parameters are sent to the cloud. Based on the distribution information provided by GMMs, the cloud picks a set of public samples that best match each device's data pattern to construct the customized metric dataset for each device, which serves as a proxy for device-specific task requirements and is subsequently used to guide the pruning process. 
Then, to achieve task-specific pruning, we propose an efficient dual-granularity importance evaluation-based pruning strategy. Based on the constructed metric dataset, we develop Composite Neuron Importance Evaluation and Adaptive Layer Importance Evaluation mechanisms to accurately estimate task-related neuron and layer importance and determine the pruning strategy accordingly. The former evaluates individual neurons using multiple criteria to decide which neurons to prune, and the latter measures each layer's task relevance to determine how aggressively that layer should be pruned. Through this two-stage importance evaluation and pruning, TAP-ViTs adaptively compresses ViTs for devices to achieve desired task-specific performance.
Extensive experiments across various ViT architectures demonstrate the effectiveness and practicability of our approach. Compared to recent state-of-the-art methods, TAP-ViTs achieves superior performance on local data with minimum overhead for the pruning process. Particularly, the customized pruned ViT even outperforms the original full ViT under low pruning ratios, highlighting its ability to eliminate redundancy while preserving task-critical model parameters. 
Our main contributions are summarized as follows: 
\begin{itemize}
\item We propose TAP-ViTs, a novel task-adaptive pruning framework for ViTs that customizes pruned ViTs for on-device deployment while requiring no direct access to device-local data. 
\item We design a GMM-based metric dataset construction method to model and approximate local data distributions and task requirements while preventing the leakage of raw local data. 
\item We develop a composite neuron and adaptive layer pruning strategy that effectively adapts pruned ViTs to device-specific tasks. 
\item Extensive experiments show that TAP-ViTs surpasses state-of-the-art pruning baselines at lower cost and even outperforms unpruned models under low pruning ratios. 
\end{itemize}

The rest of this paper is organized as follows. 
Section~\ref{sec:Related Works} reviews related work on ViT pruning and on-device deployment.
Section~\ref{sec:Preliminary} lays the groundwork by introducing ViT architectures, empirically motivating task-adaptive pruning, and formulating edge pruning as an optimization problem that preserves privacy and satisfies hardware constraints.
Section~\ref{sec:Methodology} details the proposed TAP-ViTs framework, including the GMM-based metric dataset construction and the dual-granularity importance evaluation–based pruning strategy. 
Section~\ref{sec:Experiments} presents extensive experimental results across multiple ViT architectures and datasets.
Section~\ref{sec:Conclusion and Discussion} concludes the paper and discusses future research directions. 
\section{Related Works} \label{sec:Related Works}

%In this section, we review previous studies relevant to our work, including advancements in Vision Transformers, pruning strategies tailored for these models.

\subsection{Vision Transformers}
The pioneering work of \cite{ViT} introduced Transformer architectures to computer vision by splitting images into patches and projecting them into a latent space. Vision Transformers (ViTs) capture long-range dependencies and, when trained on large-scale datasets (e.g., JFT-300M), achieve state-of-the-art performance in image classification.
DeiT introduced by Touvron et al. \cite{DeiT}, incorporated a knowledge distillation framework where a teacher network, often a convolutional neural network, guided the training process. This significantly improved data efficiency, enabling Transformer-based models to perform well even with relatively limited data. Concurrently, T2T-ViT \cite{t2t} refined patch tokenization by progressively aggregating neighboring tokens, enhancing local feature extraction while preserving global representation learning—an essential factor for effective image understanding. Building on these insights, the Swin Transformer \cite{swin} adopted a hierarchical architecture with a shifted windowing mechanism, restricting self-attention computation to local windows to reduce complexity while enabling indirect cross-window interactions to aggregate global context.
These seminal contributions have laid the foundation for the development of mainstream Vision Transformer architectures in subsequent research.
% Despite their strong performance on numerous vision tasks, the high computational and memory demands of ViTs remain a bottleneck for deployment on resource-constrained devices.
Despite their strong performance, standard ViTs impose high computational and memory demands, creating a bottleneck for deployment on resource-constrained mobile and IoT devices. To address this, recent research has explored mobile-friendly hybrid architectures. Models such as MobileViT \cite{mobilevit} and EdgeNext \cite{edgenext} combine the efficiency of lightweight CNNs (e.g., depth-wise separable convolutions) with the global context capabilities of Transformers. However, even these optimized architectures often require further compression to meet the strict latency and energy constraints of diverse edge hardware.

\subsection{Pruning in ViTs}
To mitigate the demanding resource requirements of ViTs, a range of pruning techniques have been proposed, which can be broadly categorized into weight pruning and token pruning.
Early weight pruning methods \cite{old0, old1}, such as magnitude-based pruning, remove weights with small absolute values or gradients, based on the assumption that they contribute little to the final output. Although conceptually simple, such heuristics overlook the multi-faceted functional roles of transformer neurons. Subsequent research \cite{new0, new1} has advanced this direction by introducing more sophisticated importance estimators—such as sensitivity-based analyses, dependency-aware scoring functions, and sparsity-guided search procedures—that capture substantially richer neuron characteristics beyond mere magnitude, enabling more effective preservation of model performance after pruning.
These methods generally rely on a predefined, fixed metric dataset to score neuron importance and prune a uniform fraction of neurons across layers. While effective on centralized datasets, such designs implicitly assume that all devices share similar data distributions and task characteristics, which is rarely true in heterogeneous on-device environments. As a result, the pruned model is “one-size-fits-all,” unable to adapt to device-specific task requirements. Moreover, most works adopt single-dimensional importance metrics (e.g., dependency alone), failing to characterize the inherently composite nature of neuron utility, which limits their ability to accurately capture the multifaceted contributions of neurons to task-specific performance.
Complementary to weight pruning, token pruning approaches \cite{token0, token1, token2, token3} accelerate ViTs by discarding less informative tokens based on attention entropy, predictive uncertainty, or reinforcement learning policies. These methods yield substantial speedups during inference by reducing sequence length. However, because token pruning preserves the model weights, it does not reduce parameter count, offering limited benefit for memory- or storage-constrained devices where model size is the primary bottleneck. Token pruning thus addresses computational efficiency but leaves the core challenge of model compactness unresolved. 
Nonetheless, they can serve as a complementary technique when combined with weight pruning to further optimize on-device performance \cite{all0, all1}.

\subsection{Privacy-Preserving and Task-Adaptive Deployment}
A critical challenge in mobile computing is tailoring models to user-specific data distributions without compromising privacy \cite{CJE0, CJE1}. Existing methods for personalized model compression generally fall into two categories, both of which have significant limitations for mobile deployment.
First, task-conditioned strategies \cite{Personalized0, Personalized1} attempt to adapt the pruning policy based on task identifiers or auxiliary inputs. However, these methods typically assume access to a sample of the target domain data for calibration or fine-tuning. In privacy-sensitive scenarios (e.g., smart home cameras or healthcare wearables), transmitting raw user data to a central server for calibration is prohibitive.
Second, Federated Learning (FL) frameworks have been proposed to guide pruning across clients without centralizing data \cite{fed0, fed1, fed2, fed3}. In these schemes, devices collaboratively train or prune a model by exchanging gradients or model updates. While FL preserves data privacy, it necessitates iterative on-device training or fine-tuning, which incurs substantial computational overhead and energy consumption—often impractical for low-resource edge devices with limited battery life \cite{B1,B2}. Furthermore, FL suffers from communication bottlenecks and statistical heterogeneity (Non-IID data) that can degrade the convergence of the pruned model \cite{B3,B4}.

\section{Preliminary} \label{sec:Preliminary}

\subsection{The Architecture of ViTs}

ViTs split an input image into non-overlapping patches and project them into a sequence of embeddings. A class token and positional embeddings are added to form the input sequence for the Transformer encoder. Each encoder block consists of two core modules: multi-head self-attention (MHA) and feed-forward network (FFN), both wrapped with residual connections and layer normalization. Let \( Z \in \mathbb{R}^{(N+1) \times d} \) denote the input to MHA and FFN in a given block.
In MHA, queries, keys, and values are computed via linear projections:$Q=ZW_Q, K=ZW_K, V=ZW_V.$
where \(W_Q, W_K, W_V \in \mathbb{R}^{d \times d_k}\) are learnable projection matrices and \(d_k\) denotes the dimensionality per head. The attention weights are computed as:
$
    \text{Attention}(Q, K, V) = \operatorname{softmax}\!\left( \frac{QK^\top}{\sqrt{d_k}} \right)V.
$
The outputs from multiple attention heads are concatenated and linearly projected to form the final output of the MHA.
The FFN applies two fully-connected layers with a non-linear activation (e.g., GELU) in between:
$\operatorname{FFN}(Z) = W_2 \, \sigma(W_1 Z).$
where \(W_1 \in \mathbb{R}^{d \times d'}\) and \(W_2 \in \mathbb{R}^{d' \times d}\) are learnable weight matrices, \(d'\) is the hidden dimension, and \(\sigma(\cdot)\) denotes the activation function.

\subsection{Empirical Analysis of Task-Specific Variability}

% To examine whether pruning in Vision Transformers (ViTs) should adapt to task characteristics, we conduct an empirical analysis of neuron-level importance across ten disjoint visual tasks. For each task, we compute neuron importance scores using a standard ViT and measure cross-task divergence by evaluating pairwise differences among the importance distributions. The pronounced divergence suggests that neurons critical for one task may have limited relevance to another. To further quantify this variability, we compute Kendall's~$\tau$ rank correlation coefficient between the neuron-importance rankings of different task pairs. The consistently low $\tau$ values indicate weak ordinal agreement across tasks, demonstrating that neuron saliency in ViTs is highly task-dependent. These findings reveal that a single, task-agnostic neuron-importance ranking fails to generalize across diverse task settings, underscoring the necessity of task-adaptive pruning strategies rather than uniform pruning schemes.

\begin{figure}[t]
    \centering
    \includegraphics[width=\linewidth]{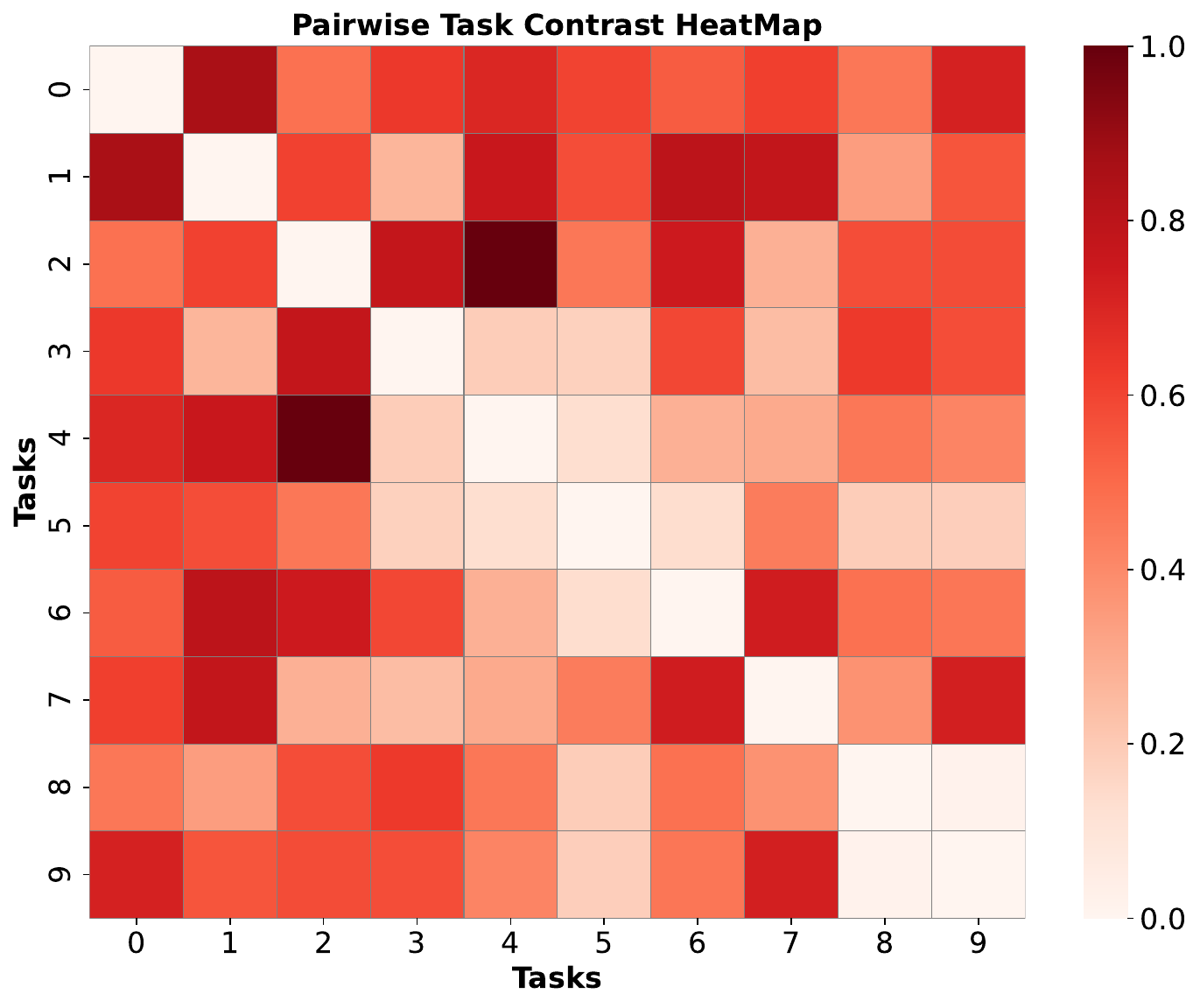}
    \caption{Heatmaps of cross-task neuron-importance divergence.}
    \label{fig:Heatmap}
\end{figure}

To assess whether pruning in Vision Transformers (ViTs) should be conditioned on task characteristics, we conduct a systematic analysis of neuron-level importance across ten heterogeneous visual tasks. For each task, we compute neuron-importance scores using a standard ViT and quantify cross-task variability by measuring the pairwise divergence between their importance distributions. As shown in Figure~\ref{fig:Heatmap}, the resulting heatmap exhibits substantial variation across task pairs, with several pairs displaying near-orthogonal importance patterns. 
This pronounced divergence indicates that neurons essential for one task often play a limited or entirely different role in another, revealing that the functional specialization of ViT neurons is highly task-dependent. Consequently, a single, task-agnostic importance ranking is insufficient to capture the distinct saliency structure required by different downstream objectives. These observations highlight a fundamental limitation of uniform pruning strategies and motivate the need for task-adaptive pruning mechanisms that explicitly account for the variability in task-specific neuron utility.

% \begin{figure*}
%   \centering
%   % \fbox{\rule[-.5cm]{0cm}{4cm} \rule[-.5cm]{4cm}{0cm}}
%   \includegraphics[width=0.98\textwidth]{fig1.pdf}
%   \caption{The Framework of TAP-ViTs.}
%   \label{fig:Framework}
% \end{figure*}

\subsection{System Model} \label{sec:System Model}
In this work, we consider a heterogeneous edge computing environment comprising a resource-rich cloud server and \( m \) end devices with limited resources and heterogeneous data characteristics. 
The cloud server, denoted by \( C \), 
% is endowed with ample computational and storage resources and 
hosts the original ViT model \( M_c \) along with a large-scale public dataset \( D_c \). 
Meanwhile, each end device \( E_i \) (with \( i \in \{1, 2, \dots, m\} \)) holds a local dataset \( D_i \) of size \(|D_i|\), and \( D_1, D_2, \dots, D_m \) are non-independent and identically distributed (Non-IID). Due to privacy concerns, these end devices are unwilling to share raw data with the cloud, thus requiring the local deployment of a ViT to process the local data.
However, their limited computational resources make deploying the full ViT \( M_c \)  infeasible. Besides, their unique local data distributions lead to diverse task requirements. 
To address this dilemma, the cloud server customizes task-adaptive pruned ViTs for on-device deployment without accessing raw local data. That is, for each resource-constrained device \( E_i \), the cloud server crafts a task-adaptive pruned ViT model \( M_i^p \) based on \( M_c \) to meet the local resource constraint and achieve high task-specific performance on \( D_i \).

\subsection{Problem Formulation} \label{sec:Problem Formulation}

The primary objective of this work is to design a pruning framework that can generate customized pruned ViTs for heterogeneous end devices without accessing their private local datasets. 
To enable the customized pruning process without accessing raw local data, we assume that for each resource-constrained device \( E_i \), there is a proxy information \( \theta_i \) that approximates the distribution of \( D_i \), which helps to perceive the individual task requirement of \( E_i \).
Then the available information for the task-adaptive pruning process is the original ViT model \( M_c \), the public dataset \( D_c \), and the proxy information of local data distribution \( \{\theta_i\}_{i=1}^m \).
Let \( \mathcal{P}(M_c, D_c, \theta_i) \) denote a pruning function that generates device-specific compressed models without accessing local raw data.

We formulate our design objective as the following optimization problem: 
\[
\begin{aligned}
\min_{M_i^p} \quad & \mathcal{L}_{D_i}(M_i^p) \\
\text{subject to} \quad & M_i^p = \mathcal{P}(M_c, D_c, \theta_i), \\
& \mathcal{R}(M_i^p) \leq \lambda_i \cdot \mathcal{R}(M_c), \\
& D_i \notin \mathcal{P},
\end{aligned}
\]
where \( \mathcal{L}_{D_i}(M_i^p) \) denotes the loss of the pruned model \( M_i^p \) on the private dataset \( D_i \),
\( \mathcal{R}(\cdot) \) represents the resource consumption required to run the model, and $0 < \lambda_i < 1$. The constraint \( D_i \notin \mathcal{P} \) ensures that the local dataset is never accessed during pruning.
With such an optimization problem, the pruning method \( \mathcal{P} \) guarantees the derivation of a device-specific pruned model \( M_i^p \), which is optimized to maximize task performance while strictly adhering to the local resource constraints, ensuring both privacy preservation and hardware-awareness.

\begin{figure*}
  \centering
  % \fbox{\rule[-.5cm]{0cm}{4cm} \rule[-.5cm]{4cm}{0cm}}
  \includegraphics[width=0.98\textwidth]{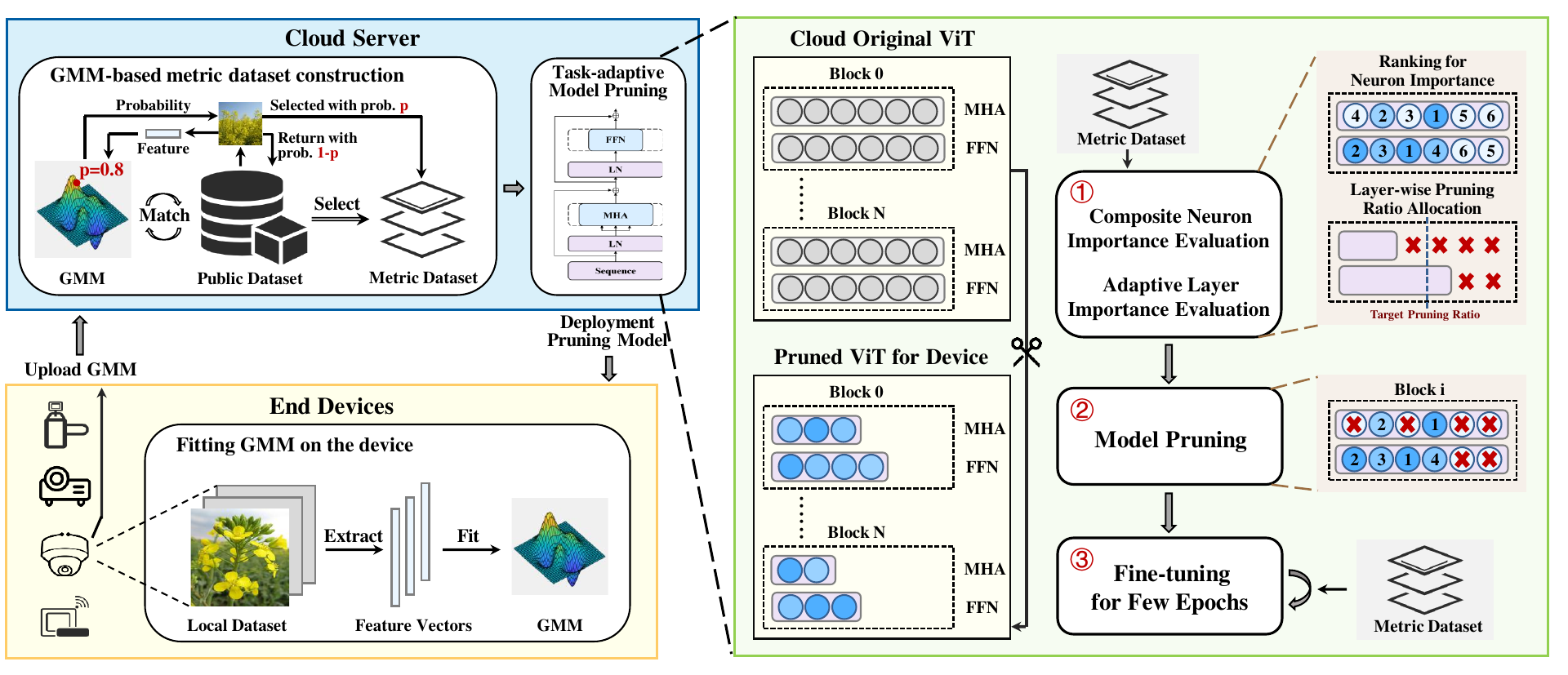}
  \caption{The Framework of TAP-ViTs.}
  \label{fig:Framework}
\end{figure*}

\section{Methodology} \label{sec:Methodology}
To solve the problem identified in ~\ref{sec:Problem Formulation}, we propose TAP-ViTs, a novel task-adaptive pruning framework tailored for ViTs. As illustrated in Figure~\ref{fig:Framework}, TAP-ViTs consists of two key components: 
1) a GMM-based metric dataset construction method. It customizes a metric dataset for each end device by approximating their local data distribution  with GMM and then selecting representative public samples accordingly. 2) an efficient task-adaptive pruning method based on comprehensive importance evaluation. It evaluates composite neuron importance and adaptive layer importance to determine both neuron-level and layer-level pruning strategies based on the metric dataset. Combining these components together, TAP-ViTs is enabled to generate customized pruned ViT models that preserve task-specific performance while ensuring data privacy for each device.

\subsection{GMM-based Metric Dataset Construction }
To enable task-adaptive pruning without accessing raw device-local data, it is necessary to construct a metric dataset that closely approximates the device's local data distribution and accurately reflects the device's individual task requirement to guide the pruning process. TAP-ViTs introduces a GMM-based mechanism for constructing a task-representative metric dataset on the cloud. The complete method is formalized in Algorithms~\ref{alg:dataset-construction}.

Let \( D_i^c \) denote the constructed metric dataset for device \( i \), it should satisfy the following objective:
\begin{equation}
%\min_{D_i^c} \mathcal{S}\left(P(D_i) \,\|\, P(D_i^c)\right),
\mathcal{D}_i^c = \arg\min_{\mathcal{D} \subseteq \mathcal{D}_c} \mathcal{S} \left( P(\mathcal{D}_i) \,\|\, P(\mathcal{D}) \right),
\end{equation}
where \( D_i \) denotes the private dataset of device \( i \), \( P(\cdot) \) represents the underlying data distribution, and \( \mathcal{S}(\cdot) \) is the Kullback-Leibler (KL) divergence measuring distributional discrepancy.

To this end, we propose a Gaussian Mixture Model (GMM)-based distribution-aligned metric dataset construction mechanism. It first estimates the latent data distribution of the device's local task via GMM. Then, leveraging the inferred distribution, it selectively retrieves a distribution-consistent subset from the publicly available cloud dataset to serve as the metric dataset to craft task-adaptive pruned ViT for the device.
To be specific, for a device with local data $D_i$, GMM models the local feature distribution as a means to estimate \( P(D_i) \).
Given a set of feature vectors \( x \in \mathbb{R}^d \), the GMM models their distribution as:
\begin{equation}
p(x) = \sum_{k=1}^{K} \pi_k \, \mathcal{N}(x \mid \mu_k, \Sigma_k),
\end{equation}
where \( K \) is the number of Gaussian components, \( \pi_k \in [0,1] \) is the mixing coefficient for the \( k \)-th component such that \( \sum_{k=1}^{K} \pi_k = 1 \), and \( \mathcal{N}(x \mid \mu_k, \Sigma_k) \) is the Gaussian density with mean \( \mu_k \in \mathbb{R}^d \) and covariance matrix \( \Sigma_k \in \mathbb{R}^{d \times d} \).
Once the GMM is fitted to the local feature representations, it serves as a compact summary of the local data distribution. 
Instead of uploading raw local data with the cloud, in TAP-ViTs, the device only needs to share the GMM parameters (\( \{\pi_k, \mu_k, \Sigma_k\}_{k=1}^{K} \)) with the cloud, thus preventing its local data from being disclosed.

Then, on the cloud side, we utilize GMMs received from devices to select representative samples from a large pool of public data to construct the customized metric dataset $\mathcal{D}_i^c = \{x_1, \dots, x_N\}$ for each device. Specifically, each public sample is first projected into the same feature space and then evaluated by its likelihood under the received GMM.
The likelihood score is computed as the log of the weighted sum of Gaussian densities defined by the GMM components, which indicates how well a public sample aligns with the statistical structure of the device-local data. Intuitively, samples with higher likelihoods are more consistent with the target device's data distribution and can contribute more to the individual task requirement. Therefore, we select public samples with the highest likelihood scores to form the metric dataset to fully reflect the task requirement of each device, providing precision guidance for the pruning process.

\begin{algorithm}
\caption{GMM-based Metric Dataset Construction}
\label{alg:dataset-construction}
\begin{algorithmic}[1]
\Require Device dataset $\mathcal{D}_i$, Public dataset $\mathcal{D}_c$, and Feature extractor $f_{ex}(\cdot)$
\Ensure Metric dataset: $\mathcal{D}_i^c$

\State \textbf{On Device Side:}
\State Extract feature representations for device data: \quad 
\[
\mathcal{F}_i \gets \{ f_{\text{ex}}(x) \mid x \in \mathcal{D}_i \},  \quad f_{\text{ex}}(x) \in \mathbb{R}^d
\]
\State Fit a GMM to local features: \quad 
% \[
$\theta_i \gets \text{GMM}(\mathcal{F}_i)$
% \]
\State Upload GMM parameters $\theta_i = \{\pi_k, \mu_k, \Sigma_k\}_{k=1}^{K}$ to cloud

\Statex

\State \textbf{On Cloud Side:}
\State Project public data into feature space: \quad 
\[
\mathcal{F}_c \gets \{ f_{\text{ex}}(x) \mid x \in \mathcal{D}_c \}, \quad f_{\text{ex}}(x) \in \mathbb{R}^d
\]
\State Evaluate sample likelihoods using $\theta_i$:
\ForAll{$x \in \mathcal{D}_c$}
    \State %Compute likelihood score:
    % \[
    $p(x) \gets \log \left( \sum_{k=1}^K \pi_k \, \mathcal{N}(f_{\text{ex}}(x) \mid \mu_k, \Sigma_k) \right)$
    % \]
\EndFor
\State Select top-N samples with highest likelihood scores:
\[
\mathcal{D}_i^c \gets \text{TopN}(\mathcal{D}_c, \{p(x)\}_{x \in \mathcal{D}_c}, N)
\]
\State \Return $\mathcal{D}_i^c$

\end{algorithmic}
\end{algorithm}

% \subsection{Importance Evaluation-based Task-Adaptive Pruning Method}
\subsection{Dual-Granularity Importance Evaluation-based Task-Adaptive Pruning}
To achieve task-specific pruning, we propose a novel dual-granularity importance evaluation-based pruning method. 
It performs neuron and layer importance evaluation based on the constructed metric dataset and determines the neuron-level and layer-level pruning strategies accordingly. Specifically, to comprehensively evaluate the importance of the model components, the method develops the following two importance evaluation mechanisms: composite neuron importance evaluation, which assesses the activeness, redundancy, and task-relevance of individual unitsand; and adaptive layer importance evaluation, which quantifies the contribution of entire layers to the model's predictive distribution. Then it adaptively allocates pruning ratios for each layer based on the pre-defined pruning ratio and layer importance and pruning neurons in each layer according to their importance scores. After that, the pruned ViTs can be fine-tuned with metric datasets to further enhance the task-specific performance. These designs enable highly customized ViT model compression that preserves task-critical capacity without accessing device-local data. The overall procedure is summarized in Algorithm~\ref{alg:task-adaptive-pruning}.

\paragraph{Composite Neuron Importance Evaluation}
To comprehensively assess the importance of each neuron, we innovatively introduce a composite evaluation strategy that integrates three complementary dimensions, i.e., activeness, redundancy, and relevance, into the estimation process. Specifically, activeness quantifies how actively a neuron responds to inputs, reflecting its contribution to feature encoding. Redundancy measures the mutual information between a neuron and its peers within the same layer, indicating the degree of functional overlap. Relevance reflects a neuron's task specificity by capturing its statistical correlation with the model's final prediction.
Formally, for a given neuron $j$, we compute the three indicators over a metric dataset $\mathcal{D}_i^c = \{x_1, \dots, x_N\}$ as follows:
\begin{subequations}
\begin{align}
A_j &= \frac{1}{N} \sum_{n=1}^{N} |a_j(x_n)|, \\
R_j &= -\frac{1}{N} \sum_{n=1}^{N} \sum_{\substack{r=1 \\ r \neq j}}^{l} 
       \frac{\text{mi}(o_j(x_n), o_r(x_n))}{l - 1}, \\
T_j &= \frac{1}{N} \sum_{n=1}^{N} \left\| H_{y_j,Y} \right\|_{\text{HS}}^2.
\end{align}
\end{subequations}
where $a_j(\cdot)$ is the activation value, $\text{mi}(\cdot,\cdot)$ is mutual information, $o_j(\cdot)$ is neuron output, $l$ is the number of neurons in the layer, and $H_{y_j,Y}$ is the RKHS cross-covariance between the neuron's output and the model output. All three indicators are normalized to ensure comparable scales.
The final importance score is then computed as a weighted sum:
\begin{equation}
I_j = \alpha A_j + \beta R_j + \gamma T_j,
\end{equation}
where $\alpha, \beta, \gamma $ denote weighting factors that balance the contributions of the three indicators. By jointly modeling activeness, redundancy, and relevance, our method enables a multi-dimensional and precise evaluation of neuron importance, offering a robust foundation for task-adaptive pruning.

\paragraph{Adaptive Layer Importance Evaluation} 
Unlike conventional pruning methods that either apply a uniform pruning ratio across all layers or rely on manually tuned layer-wise budgets, we introduce an adaptive layer importance evaluation strategy driven by the metric dataset. This enables automatic and task-specific allocation of pruning ratios based on each layer's contribution to the model's predictive behavior. Specifically, given a target global pruning ratio, we estimate the importance of each layer \(l\) by computing the average Kullback-Leibler (KL) divergence $\delta_l^{'}$ between the original model output $p(y \mid x)$ and the output after pruning layer \(l\), denoted $q_l(y \mid x)$:
\begin{equation}
\delta_l^{'} = \frac{1}{N} \sum_{n=1}^{N} S(p(y \mid x_n) \| q_l(y \mid x_n)),
\end{equation}
where \( \mathcal{S}(\cdot) \) is the KL divergence. The resulting KL-based scores are then normalized via a softmax function to determine the relative pruning budgets across layers:
\begin{equation}
\delta_l = \frac{\exp(\delta_l^{'})}{\sum_{u=1}^{L} \exp(\delta_u^{'})},
\end{equation}
A larger divergence indicates that pruning the layer leads to a more significant shift in model predictions, thus implying higher importance. This score determines the pruning budget allocated to each layer, allowing more critical layers to retain more capacity.

\begin{algorithm}
\caption{Dual-Granularity Importance Evaluation-based Task-Adaptive Pruning}
\label{alg:task-adaptive-pruning}
\begin{algorithmic}[1]
\Require Original ViT model \( M_c \), Metric dataset \( \mathcal{D}_i^c \), Target global pruning ratio \( \epsilon_t \)
\Ensure Pruned model \( M_i^p \)

\Statex

\State \textbf{Step 1: Estimate neuron-level importance scores $\{I_j\}$}
\For{each neuron \( j \) in all layers}
    \State \textbf{Activeness:}
    % \[
    $A_j = \frac{1}{N} \sum_{n=1}^{N} |a_j(x_n)|$
    % \]
    \State \textbf{Redundancy:}
    % \[
    $R_j = \frac{-1}{N} \sum_{n=1}^{N} \sum_{\substack{r=1 \\ r \neq j}}^{l} \frac{\text{mi}(o_j(x_n), o_r(x_n))}{l - 1}$
    % \]
    \State \textbf{Relevance:}
    % \[
    $T_j = \frac{1}{N} \sum_{n=1}^{N} \|H_{y_j,Y}\|_{\mathrm{HS}}^2$
    % \]
    \State Normalize \( A_j, R_j, T_j \) to comparable scales
    \State Compute final importance score:
    % \[
    $I_j = \alpha A_j + \beta R_j + \gamma T_j$
    % \]
\EndFor

\Statex

\State \textbf{Step 2: Estimate layer-level importance scores $\{\delta_l\}$}
\For{each layer \( l = 1, \dots, |L| \)}   
    \State Compute original model output \( p(y \mid x_n) \)
    \State Temporarily disable layer \( l \), get modified model \( M_c^{(-l)} \)
    \State Compute modified output \( q_l(y \mid x_n) \)
    \State Compute KL divergence between outputs:
    \[
    \delta_l' = \frac{1}{N} \sum_{n=1}^{N} \mathcal{S}(p(y \mid x_n) \,\|\, q_l(y \mid x_n))
    \]
\EndFor
\State Normalize layer importance score:
% \[
$\delta_l = \frac{\exp(\delta_l')}{\sum_{u=1}^{|L|} \exp(\delta_u')}$
% \]

\Statex

\State \textbf{Step 3: Adaptive Pruning and Few-shot Fine-tuning}
\For{each layer \( l = 1, \dots, |L| \)}
    \State Compute layer-specific pruning ratio: 
    % \[
    $\epsilon_l = \delta_l \cdot |L| \cdot \epsilon_t$
    % \]
    \State Rank neurons in layer \( l \) by neuron importance \( \{I_j\} \)
    \State Prune the bottom \( \epsilon_l \times 100\% \) neurons with lowest \(I_j\)
\EndFor
\State Obtain pruned model \( M_i^p \) by removing all pruned neurons
\State Fine-tune \( M_i^p \) on \( \mathcal{D}_i^c \) for a few epochs to recover task performance

\Statex

\State \Return \( M_i^p \)
\end{algorithmic}
\end{algorithm}

% \paragraph{Pruning and Fine-tuning.} Based on the neuron-level importance scores $I_j$ and the normalized layer-level importance scores $\delta_l$, we devise a pruning scheme that preserves high-importance neurons while adhering to a global pruning ratio. For each layer, neurons with the lowest importance scores are pruned according to an allocated budget.
% To achieve a target overall pruning ratio denoted by \(\epsilon_t\), we first determine the pruning rate for each layer. Since the layer importance scores have been normalized, the pruning ratio for layer \(l\), denoted by \(\epsilon_l\), is computed as:
% \begin{equation}
%     \epsilon_l = \delta_l \times |L| \times \epsilon_t,
% \end{equation}
% where \(\delta_l\) is the normalized importance score of layer \(l\) and \(|L|\) is the total number of layers.
% We then prune the lowest-ranking neurons from each layer according to their respective \(\epsilon_l\), effectively distributing the pruning budget in proportion to layer-wise importance. To mitigate potential performance degradation introduced by pruning, we apply a lightweight few-shot fine-tuning step using the metric dataset.
\paragraph{Pruning and Fine-tuning.} 
Guided by the neuron-level importance scores $I_j$ and the normalized layer-level importance scores $\delta_l$, we design a global pruning strategy that allocates pruning budgets across layers in proportion to their relative importance. The goal is to preserve highly informative neurons while achieving a target global pruning ratio $\epsilon_t$.
To determine the pruning budget for each layer, we compute a layer-wise pruning ratio $\epsilon_l$ based on its normalized importance value:
\begin{equation}
    \epsilon_l = \delta_l \times |L| \times \epsilon_t,
\end{equation}
where $|L|$ denotes the total number of layers. Since the layer-importance scores are normalized such that $\sum_l \delta_l = 1$, this formulation guarantees that the total pruning budget satisfies the global pruning constraint. 
In particular,
% \[
$\sum_{l=1}^{|L|} \epsilon_l 
= \sum_{l=1}^{|L|} \delta_l \cdot |L| \cdot \epsilon_t
= |L| \cdot \epsilon_t,$
% \]
which ensures that the average pruning ratio across layers is exactly $\epsilon_t$. Thus, although pruning is applied independently to each layer, the global pruning requirement is met by distributing budgets proportionally to layer importance.
After computing $\epsilon_l$, we prune the lowest-ranked neurons within each layer according to its assigned budget. To mitigate potential performance degradation caused by pruning, we further apply a lightweight few-shot fine-tuning step using the metric dataset, enabling the pruned model to recover essential task-relevant representations without accessing device-local data.

\section{Experiments} \label{sec:Experiments}
% In this section, we present comprehensive experimental results to demonstrate the effectiveness and generality of our proposed \textbf{TAP-ViTs} framework. We evaluate our method across multiple Vision Transformer architectures and compare TAP-ViTs against several state-of-the-art pruning baselines In addition to performance comparisons, we perform ablation studies to assess the effect of the constructed metric dataset and the pruning strategy based on composite neuron and adaptive layer importance. Finally, we validate the real-world practicality of TAP-ViTs through deployment experiments on a testbed edge device, focusing on the time required to model local data distributions. Notably, this is the only on-device step required throughout the entire pruning process, highlighting the negligible computational burden imposed on edge devices.
In this section, we conduct a comprehensive empirical study to evaluate the effectiveness, robustness, and deployment practicality of TAP-ViTs. Our experiments primarily focus on image classification across diverse Transformer architectures and datasets, where we benchmark TAP-ViTs against strong state-of-the-art pruning baselines under various fine-tuning budgets and compression ratios. To better contextualize the necessity of task-adaptive pruning, we first introduce a dedicated analysis on task sensitivity in ViTs. We further investigate the behavior of our method through detailed ablations that isolate the contribution of each design component and analyze the sensitivity of TAP-ViTs to key algorithmic choices. 
% To demonstrate task generality beyond classification, we additionally report a small-scale object detection experiment using a pruned PVT-Small backbone in RetinaNet, serving as a lightweight supplementary validation. 
Finally, we assess real-world deployability by profiling the end-to-end cost of the lightweight GMM fitting performed on-device within practical deployment settings.

\subsection{Experimental Setup} \label{sec:Setup}

% \textbf{Models and Datasets.} We conduct experiments on two standard ViT models: DeiT-Base\cite{DeiT} and DeiT-Small\cite{DeiT}, as well as two popular ViT variants: T2T-ViT\cite{t2t} and Swin Transformer\cite{swin}. Two widely used datasets are used: TinyImageNet\cite{imagenet} and CIFAR-100\cite{cifar100}. The whole test set is partitioned and distributed across multiple devices. Each device evaluates the pruned model on its assigned subset.
\subsubsection{Models and Datasets} We conduct experiments on two standard ViT models, DeiT-Base~\cite{DeiT} and DeiT-Small~\cite{DeiT}, as well as two representative ViT variants, T2T-ViT~\cite{t2t} and Swin Transformer~\cite{swin}. Experiments are performed on two widely used datasets: TinyImageNet~\cite{imagenet} and CIFAR-100~\cite{cifar100}. To emulate realistic device-specific task distributions, we partition the full test split of each dataset by semantic labels. The label space is divided into ten disjoint groups, and each device is allocated the subset of images associated with its assigned label group. This setup ensures that each device evaluates the pruned model on a class-specific test subset, mirroring practical deployment scenarios in which heterogeneous devices encounter distinct portions of the visual category space.

\subsubsection{Compared Methods} 
% We compare TAP-ViTs with three strong baseline methods: \textbf{MD-ViT}\cite{mdvit}, \textbf{UP-ViT}\cite{upvit}, and \textbf{DC-ViT}\cite{dcvit}.

We compare TAP-ViTs with three strong and representative ViT compression baselines: \textbf{MD-ViT}\cite{mdvit}, \textbf{UP-ViT}\cite{upvit}, and \textbf{DC-ViT}\cite{dcvit}. 
MD-ViT adopts a multi-dimensional pruning strategy that jointly compresses attention heads, feedforward neurons, and token sequences, guided by a dependency-based importance criterion and optimized using a Gaussian Process search. UP-ViT provides a unified structured pruning framework that estimates channel importance across MHA, FFN, normalization, and further integrates progressive block pruning to enhance architectural diversity. DC-ViT targets data-constrained settings by performing fine-grained structural modification, selectively pruning attention modules while reusing MLP components to construct dense model variants across a wide range of computation budgets. Together, these baselines represent state-of-the-art approaches to ViT compression under diverse assumptions regarding data availability and structural flexibility.

% \subsubsection{Pruning Settings} In the main comparison, we prune all models to retain \textbf{70\%} of the original parameters and fine-tune the pruned models using either \textbf{1/10} of TinyImageNet or \textbf{1/5} of CIFAR-100 for only \textbf{50 epochs}. For a comprehensive evaluation, we also include a more relaxed setting where baseline methods are fine-tuned using the \textbf{full} training set for \textbf{200 epochs}. 
% All hyperparameters used in our experiments are selected via grid search on a validation set and fixed during pruning to ensure consistent performance. 
\subsubsection{Pruning Settings} In our main comparison, we prune all models to retain \textbf{70\%} of the original parameters. Following pruning, all baselines and our method undergo lightweight fine-tuning using either \textbf{1/10} of TinyImageNet or \textbf{1/5} of CIFAR-100 for only \textbf{50 epochs}. For a comprehensive evaluation, we additionally consider a more relaxed setting in which baseline pruning methods are afforded access to the \textbf{full} training set and are fine-tuned for \textbf{200 epochs}. This extended regime serves as an upper bound on the performance that existing approaches can achieve when unconstrained by data availability or computational budget. Comparing results across the two regimes highlights the robustness of our method and its ability to operate effectively under realistic on-device constraints.

\subsubsection{Hyperparameters Settings}
We further provide details on the hyperparameters involved in our pruning framework.

\textbf{Composite Scoring Weights}. Our neuron-importance estimator integrates three complementary criteria---activeness ($A_j$), redundancy ($R_j$), and relevance ($T_j$)---via a composite score $I_j=\alpha A_j + \beta R_j + \gamma T_j$. The weighting coefficients $(\alpha, \beta, \gamma)$ critically influence pruning decisions. To determine appropriate values, we perform a grid search over the 3-simplex with a step size of 0.1, yielding 66 valid configurations that satisfy $\alpha+\beta+\gamma=1$. For each configuration, we prune the model and select the triplet achieving the highest validation accuracy. The optimal configuration is found to be $(\alpha, \beta, \gamma)=(0.1, 0.1, 0.8)$, indicating that task relevance plays a dominant role while modest contributions from activeness and redundancy improve robustness. These weights are fixed throughout all experiments for consistency across pruning stages.

\textbf{Choice of GMM Components}. For estimating local feature distributions, we adopt a Gaussian Mixture Model (GMM). The number of components $K$ significantly impacts both model fidelity and computational cost. We select $K$ in a data-driven manner using the Bayesian Information Criterion (BIC). For each device, we evaluate GMMs with $K\in\{2,\dots,10\}$ and choose the model that minimizes the BIC. This procedure ensures that the selected GMM adequately captures the complexity of the device-specific feature distribution while avoiding unnecessary over-parameterization. Empirically, we observe that the optimal $K$ varies across devices, reflecting the heterogeneous nature of class-specific tasks in edge deployment.

\subsubsection{Metric} 
% As our pruning strategy is designed for task-specific end deployment, test data varies across devices. The full test set is partitioned over 10 end devices, and pruning accuracy is reported as a \textbf{weighted average of device-specific accuracies}, based on each device's sample proportion.
Since our pruning strategy targets task-specific on-device deployment, the evaluation data naturally differs across devices. As described earlier, the full test split is partitioned into ten disjoint, label-based subsets, each representing the class distribution observed by a particular end device. To provide a unified evaluation across this heterogeneous setting, we report the overall pruning accuracy as a \textbf{weighted average of device-specific accuracies}, where weights correspond to the relative number of test samples allocated to each device. This metric captures the aggregated performance under realistic, non-overlapping device workloads and reflects the effectiveness of the pruned model in practical multi-device deployment scenarios.

\subsubsection{Experimental Environment} All experiments are conducted with PyTorch on a single NVIDIA RTX 4090 GPU for consistency and reproducibility. To further evaluate on-device efficiency, we perform deployment experiments on a real-world testbed consisting of NVIDIA Jetson Orin Nano 8GB.
% The device serves as a representative edge platform with constrained computational resources, enabling us to measure actual end-to-end latency, throughput, and memory usage. These on-device results complement the GPU-based evaluation and demonstrate the applicability of TAP-ViTs to resource-limited deployment settings.
The device serve as representative edge platforms with limited computational resources, allowing us to measure the actual end-to-end latency of the only device-side computation required by our framework: GMM fitting to approximate the local data distribution. 
The results provide a realistic assessment of TAP-ViTs' scalability and practical feasibility under constrained computational conditions, complementing the GPU-based evaluation.

% More details can be found in the Appendix of the supplementary material.

% All experiments are conducted under the PyTorch framework using a single NVIDIA RTX 4090 GPU to ensure consistency and reproducibility. Besides, to evaluate on-device efficiency, we perform deployment experiments on a real-world testbed, which consists of NVIDIA Jetson Orin Nano 8GB.

\begin{table*}[t]
\centering
\tiny
\caption{Comparison of TAP-ViTs Versus Baselines on TinyImagenet and CIFAR-100}
\label{tab:comparison}
\resizebox{\textwidth}{!}{
\begin{tabular}{llccccccccccccc}
\toprule
\multirow{2}{*}{\textbf{Model}} 
& \multirow{2}{*}{\textbf{Methods}} 
& \multirow{2}{*}{\textbf{Param / Retention Rate}} 
& \multicolumn{2}{c}{\textbf{TinyImageNet}} 
& \multicolumn{2}{c}{\textbf{CIFAR-100}} \\
\cmidrule(lr){4-5} \cmidrule(lr){6-7}
& & 
& \textbf{Acc1 (\%)} 
& \textbf{Acc5 (\%)} 
& \textbf{Acc1 (\%)} 
& \textbf{Acc5 (\%)} \\
\midrule
\multirow{7}{*}{DeiT-B}
 & Original & 86M ~/~ 100\% & 85.38 & 96.08 & 90.02 & 98.71 \\
 & \textbf{Ours} & 60.2M ~/~ 70\% & \textbf{88.24} & \textbf{97.06} & \textbf{92.72} & \textbf{99.25} \\
 & MD-VIT & 60.2M ~/~ 70\% & 77.78 & 93.66 & 81.12 & 97.09 \\
 & MD-VIT\textsuperscript{*} & 60.2M ~/~ 70\% & 84.98 & 96.24 & 87.52 & 98.19 \\
 & UP-VIT & 60.2M ~/~ 70\% & 82.24 & 94.96 & 81.45 & 97.22 \\
 & UP-VIT\textsuperscript{*} & 60.2M ~/~ 70\% & 87.29 & 97.10 & 88.44 & 98.35 \\
 & DC-VIT\textsuperscript{\dag} & 60.2M ~/~ 70\% & 84.99 & 95.76 & 87.60 & 98.29 \\
\midrule
\multirow{7}{*}{DeiT-S} 
 & Original & 22M ~/~ 100\% & 79.93 & 93.76 & 87.55 & 97.73 \\
 & \textbf{Ours} & 15.4M ~/~ 70\% & \textbf{81.50} & \textbf{94.58} & \textbf{90.32} & \textbf{98.61} \\
 & MD-VIT & 15.4M ~/~ 70\% & 71.16 & 90.20 & 75.38 & 95.24 \\
 & MD-VIT\textsuperscript{*} & 15.4M ~/~ 70\% & 79.89 & 94.27 & 84.70 & 97.32 \\
 & UP-VIT & 15.4M ~/~ 70\% & 74.98 & 91.85 & 76.09 & 95.47 \\
 & UP-VIT\textsuperscript{*} & 15.4M ~/~ 70\% & 81.39 & 94.94 & 85.23 & 97.41 \\
 & DC-VIT\textsuperscript{\dag} & 15.4M ~/~ 70\% & 41.67 & 68.67 & 73.18 & 92.89 \\
\bottomrule
% \vspace{-10pt}
\end{tabular}
}
\begin{flushleft}
\footnotesize{ \ *\ means fine-tuned with full training set for 200 epochs and {\dag} means following the original paper's settings}
\end{flushleft}
% \vspace{-15pt}
\end{table*}

\subsection{Experimental Results}

\subsubsection{Comparison with Baselines}

Table~\ref{tab:comparison} summarizes the comparative performance of our proposed TAP-ViTs framework against several state-of-the-art pruning baselines on TinyImageNet and CIFAR-100, using both DeiT-Base and DeiT-Small as backbone models. To ensure a fair and controlled comparison, all pruning methods preserve exactly 70\% of the original model parameters. Notably, TAP-ViTs consistently achieves superior Top-1 accuracy across both datasets and model scales, despite being fine-tuned with smaller datasets and fewer epochs.

On the DeiT-Base architecture, TAP-ViTs achieves 88.24\% Top-1 accuracy on TinyImageNet and 92.72\% on CIFAR-100, outperforming not only all competing pruning baselines but also the original uncompressed model. Specifically, TAP-ViTs improves over the unpruned DeiT-Base by +2.86\% on TinyImageNet and +2.70\% on CIFAR-100, even though it retains only 70\% of the parameters. This trend persists when compared to strong pruning baselines such as MD-ViT, UP-ViT, and DC-ViT, all of which exhibit notable accuracy degradation under the 70\% retention setting. In contrast, TAP-ViTs maintains high stability and consistently superior accuracy across datasets.
A similar pattern emerges on the DeiT-Small architecture. TAP-ViTs achieves 81.50\% Top-1 accuracy on TinyImageNet and 90.32\% on CIFAR-100—again surpassing the original unpruned model while using 30\% fewer parameters. The relative improvements over the full model are substantial, highlighting the robustness of our pruning strategy even when applied to compact Transformers.

In addition to outperforming standard pruning baselines, TAP-ViTs also exceed several enhanced baseline variants (denoted with “*”), which undergoes substantially more aggressive fine-tuning using the full training set for 200 epochs. That TAP-ViTs matches or surpasses these heavily tuned baselines—despite employing a markedly smaller fine-tuning dataset and fewer fine-tuning epochs—demonstrates the strength of our GMM-based metric dataset construction as well as the task-aware pruning strategy, both of which jointly enable more effective parameter allocation.

Overall, these results verify that TAP-ViTs is highly task-adaptive, data-efficient, and robust under limited training budgets. Even when fine-tuning is restricted to reduced subsets of data and shorter training schedules, TAP-ViTs not only preserves the capability of the original model but frequently delivers beyond-baseline performance, setting a new benchmark for practical, task-specific pruning of Vision Transformers.

\begin{table}[t]
\centering
\caption{Performance of TAP-ViTs on T2T and Swin}
\label{tab:variants}
\renewcommand{\arraystretch}{1.15} % 稍微增加行距
\setlength{\tabcolsep}{3.5pt}      % 缩小列间距，适应单栏宽度
\resizebox{\columnwidth}{!}
{ 
\begin{tabular}{llcccc}
\toprule
\multirow{2}{*}{\textbf{Model}} & \multirow{2}{*}{\textbf{Method}} &
\multicolumn{2}{c}{\textbf{TinyImageNet}} & \multicolumn{2}{c}{\textbf{CIFAR-100}} \\
\cmidrule(lr){3-4} \cmidrule(lr){5-6}
 & & \textbf{Acc1 (\%)} & \textbf{Drop (\%)} & \textbf{Acc1 (\%)} & \textbf{Drop (\%)} \\
\midrule
\multirow{3}{*}{T2T}
 & Original (100\%) & 82.36 &  --   & 85.88 & -- \\
 & Pruned (70\%)    & 82.14 &  0.22 & 88.69 & -2.81 \\
 & Pruned (65\%)    & 79.97 &  2.39 & 88.52 & -2.64 \\
\midrule
\multirow{3}{*}{Swin}
 & Original (100\%) & 85.64 &  --   & 88.64 & -- \\
 & Pruned (70\%)    & 86.42 & -0.78 & 91.91 & -3.26 \\
 & Pruned (65\%)    & 85.19 &  0.45 & 91.58 & -2.94 \\
\bottomrule
\end{tabular}
}
\end{table}

\subsubsection{Performance on ViT Variants}
% To demonstrate the broad applicability of TAP-ViTs, we evaluate its performance on two representative Vision Transformer architectures—T2T-ViT and Swin Transformer—using the TinyImageNet and CIFAR-100 datasets. Table~\ref{tab:variants} reports the Top-1 accuracy and corresponding accuracy drop under different pruning ratios. Notably, TAP-ViTs retains high accuracy even under aggressive pruning, highlighting its ability to preserve task-critical information without accessing local data.

% On TinyImageNet, TAP-ViTs introduces minimal performance degradation. For T2T-ViT, pruning 30\% of the parameters results in only a 0.22\% accuracy drop. For Swin Transformer, TAP-ViTs surprisingly improves accuracy by 0.78\%, potentially due to the removal of redundant parameters. Even at 35\% pruning, accuracy remains competitive, demonstrating the robustness of our comprehensive importance evaluation. And the similar trends are observed on CIFAR-100. These results confirm that TAP-ViTs not only maintains performance under high compression but can even improve generalization in certain scenarios.

To further assess the generality and architectural robustness of TAP-ViTs, we extend our evaluation to two representative ViT variants—T2T-ViT and Swin Transformer—on both TinyImageNet and CIFAR-100. As shown in Table~\ref{tab:variants}, we report the Top-1 accuracy and the corresponding performance drop under different pruning ratios. These architectures differ substantially in design philosophy (token-to-token aggregation vs.\ hierarchical spatial reduction), providing a comprehensive testbed for examining the universality of our task-adaptive pruning framework.

Across both architectures and datasets, TAP-ViTs exhibits strong resilience to pruning, consistently maintaining accuracy even under aggressive parameter reduction. On TinyImageNet, pruning T2T-ViT to 70\% of its original size results in a negligible accuracy drop of only 0.22\%, indicating that TAP-ViTs effectively preserves task-relevant structures. Interestingly, for the Swin Transformer, pruning 30\% of parameters leads to a \emph{performance gain} of 0.78\%, suggesting that removing redundant or noisy components can enhance generalization. Even at a more stringent 65\% retention ratio, performance remains competitive for both models, reflecting the robustness of our GMM-driven importance estimation and pruning strategy.
A similar trend is observed on CIFAR-100. TAP-ViTs consistently achieves higher accuracy than the original T2T-ViT and Swin models across a range of pruning ratios, demonstrating improved generalization despite the reduced parameter count. These results collectively illustrate that TAP-ViTs generalizes beyond DeiT-based architectures and remains effective across diverse design paradigms. 

In summary, the results on these ViT variants highlight the versatility and effectiveness of TAP-ViTs. Our task-adaptive pruning framework not only preserves critical model structures but can also leverages architectural redundancies to improve model performance. The consistent performance across distinct design paradigms—ranging from token-to-token aggregation in T2T-ViT to hierarchical spatial modeling in Swin Transformer—demonstrates that TAP-ViTs provides a robust and broadly applicable pruning solution. These findings underscore the potential of task-adaptive pruning to facilitate efficient deployment of diverse transformer models without compromising accuracy.

\subsubsection{Impact of Parameter Retention Ratio} 
% We investigate how different parameter retention ratios affect the final performance of TAP-ViTs.
% The accuracy of TAP-ViTs with parameter retention ratios of 90\%, 80\%, 70\%, and 60\% are shown in Figure~\ref{fig:ablation-retention}. We can observe that smaller parameter retention ratios (i.e., higher pruning rates) lead to gradual degradation in accuracy, however, this degradation remains relatively minor. The overall accuracy is consistently maintained at a high level even under aggressive pruning ratios, indicating the robustness of TAP-ViTs under different compression rates. 

\begin{figure*}[t]
    \centering
    \captionsetup{skip=3pt}
    \includegraphics[width=0.92\linewidth]{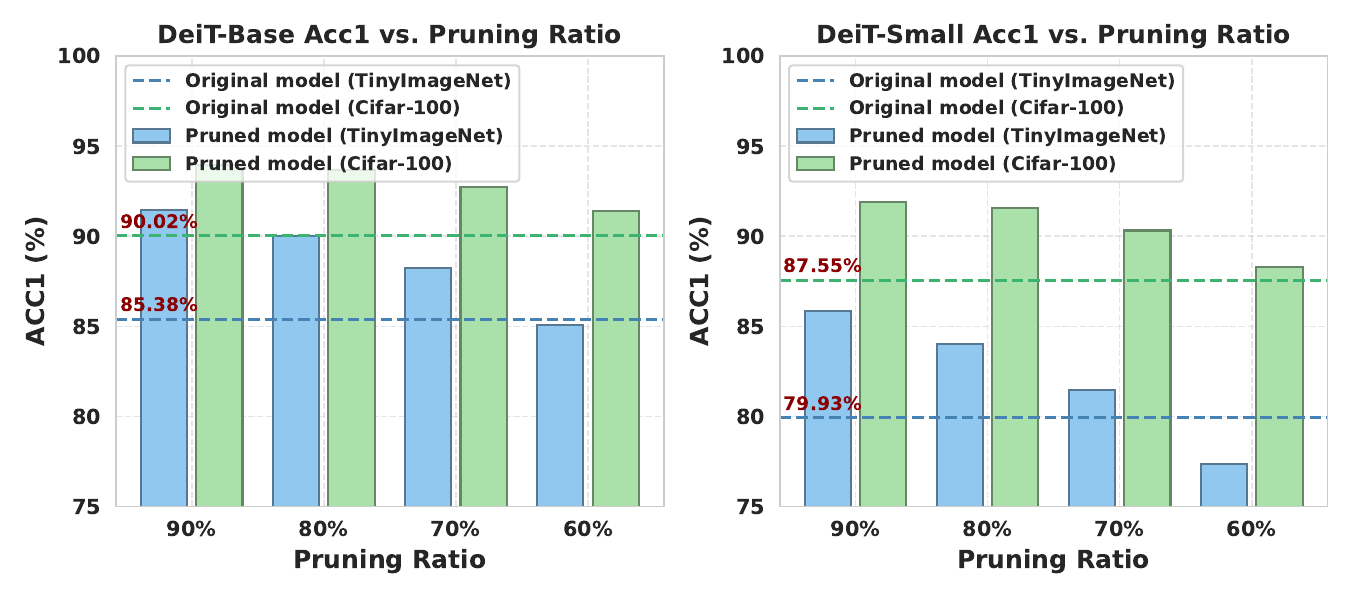}
    \caption{Performance vs. Pruning Ratio}
    \label{fig:ablation-retention}
\end{figure*}

To further understand how compression intensity influences the effectiveness of TAP-ViTs, we evaluate model performance under four parameter retention ratios: 90\%, 80\%, 70\%, and 60\%. The results for both DeiT-Base and DeiT-Small on TinyImageNet and CIFAR-100 are illustrated in Figure~\ref{fig:ablation-retention}.Overall, TAP-ViTs exhibits a clear and consistent trend: reducing the retention ratio leads to a gradual decline in accuracy, yet the degradation remains modest even under aggressive pruning. 

For DeiT-Base, TAP-ViTs maintains accuracy levels close to—or even surpassing—the original unpruned model at retention ratios as low as 60\%. Notably, on CIFAR-100, TAP-ViTs at 60\% retention still exceeds the full model’s performance, demonstrating that the proposed pruning strategy effectively removes redundant parameters while preserving task-relevant information.
A similar pattern is observed for DeiT-Small. Although accuracy decreases as compression increases, the magnitude of this drop is surprisingly small. Even at a retention ratio of 60\%, TAP-ViTs preserves competitive performance across both datasets. In particular, CIFAR-100 accuracy remains substantially above the uncompressed baseline, highlighting that smaller models benefit from our task-adaptive pruning, which acts as a form of structural regularization.

These results collectively show that TAP-ViTs provides robust performance across a wide range of compression levels. The framework gracefully trades off between compactness and accuracy, ensuring that even under high pruning ratios, the pruned models remain strong performers. This stability across retention ratios underscores the practical value of TAP-ViTs for deployment scenarios with diverse computational budgets.

\begin{table}[t]
\centering
\caption{Ablation on TAP-ViTs Components}
\label{tab:ablation-components}
\renewcommand{\arraystretch}{1.19} % 调整行距
\setlength{\tabcolsep}{5pt}        % 调整列间距

\begin{tabular}{llcc}
\toprule
\textbf{Model} & \textbf{Ablation Variant} & \textbf{TinyImageNet} & \textbf{CIFAR-100} \\
% \cmidrule(lr){3-4}
 & & \textbf{Acc1 (\%)} & \textbf{Acc1 (\%)} \\
\midrule
\multirow{3}{*}{DeiT-B}
 & Full TAP-ViTs & \textbf{88.24} & \textbf{92.72} \\
 & w/o Dataset Construction & 81.12 & 81.98 \\
 & w/o Efficient Pruning & 87.63 & 92.65 \\
\midrule
\multirow{3}{*}{DeiT-S}
 & Full TAP-ViTs & \textbf{81.50} & \textbf{90.32} \\
 & w/o Dataset Construction & 72.52 & 76.39 \\
 & w/o Efficient Pruning & 79.49 & 89.40 \\
\bottomrule
\end{tabular}
\end{table}

\subsubsection{Task Sensitivity Analysis of ViT Components}
\label{sec:task_sensitivity}

To better understand the task-dependent behavior of Vision Transformers and to justify the need for adaptive pruning, we conduct a detailed analysis of neuron- and layer-level sensitivity across diverse tasks. The goal is to answer a fundamental question: \emph{Do different tasks rely on distinct subsets of ViT neurons and layers, and if so, how substantial is this variability?}

\begin{figure}[t]
  \centering
  \includegraphics[width=\columnwidth]{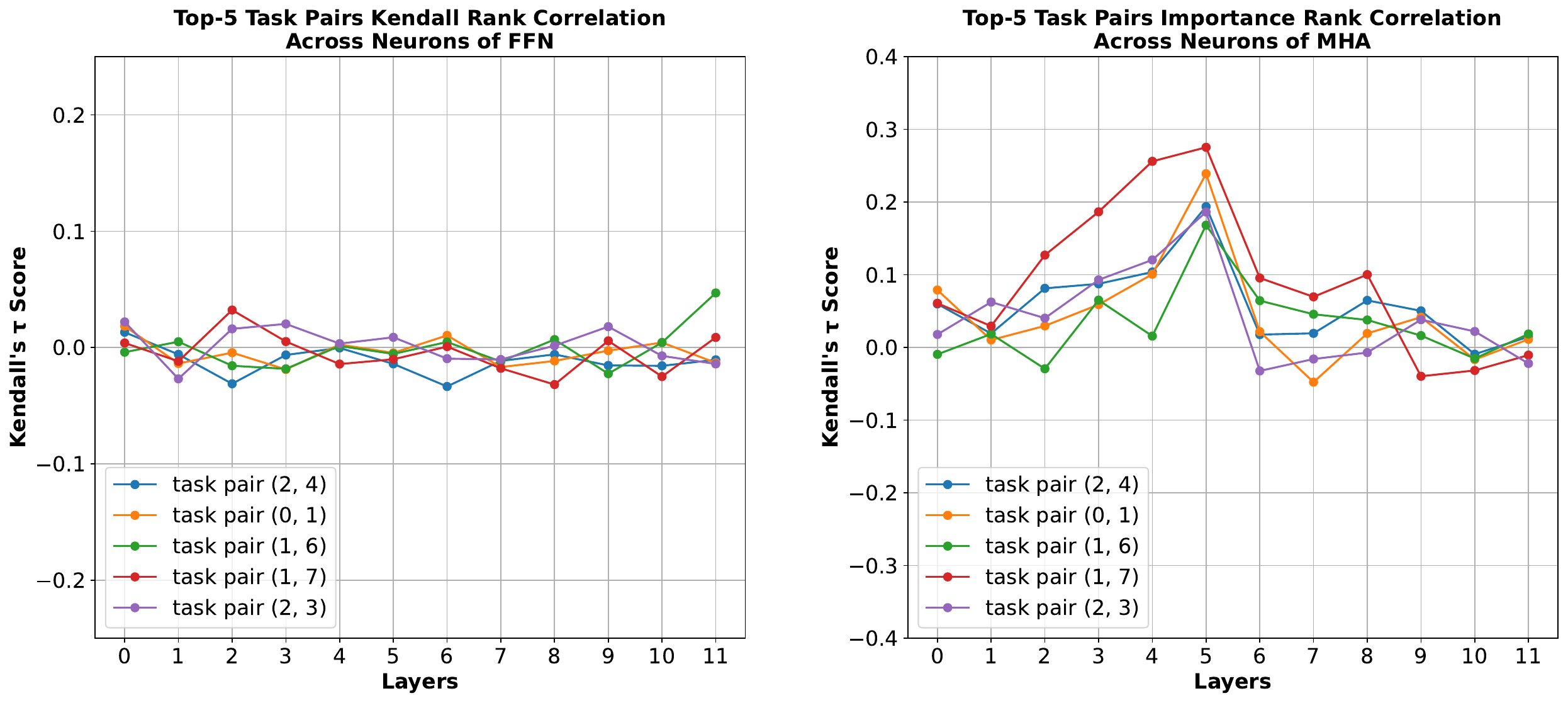}
  \caption{Kendall's $\tau$ correlation of neuron-importance rankings across tasks.}
  \label{fig:Neuron}
\end{figure}

Neuron-Level Task Sensitivity.
We begin by assessing whether neuron importance exhibits significant variability across tasks. Using a standard ViT model, we compute neuron-importance scores for ten disjoint task-specific subsets derived from the dataset. For every pair of tasks, we directly evaluate the similarity of their neuron-importance rankings using Kendall’s~$\tau$ rank correlation coefficient.
Kendall’s~$\tau$ measures the ordinal agreement between two ranked lists by comparing the number of concordant and discordant element pairs, and is defined as:
$\tau = \frac{2(P_c - P_d)}{n(n-1)},$
where $P_c$ and $P_d$ denote the numbers of concordant and discordant pairs, respectively, and $n$ is the number of ranked neurons. A value of $1$ indicates perfect agreement, $0$ indicates no correlation, and $-1$ reflects complete disagreement.
Our experimental results show that the resulting $\tau$ values across all task pairs are consistently low, indicating weak alignment in neuron-importance orderings. We further identify the five task pairs with the largest overall divergence, as illustrated in Figure~\ref{fig:Neuron}, and analyze FFN and MHA neurons separately across layers. The correlations remain weak across most layers, reaffirming that neuron relevance is highly task-dependent. These findings highlight that a single global ranking is insufficient for guiding pruning across heterogeneous tasks, thereby motivating the task-adaptive design of our pruning framework.

\begin{figure}[t]
  \centering
  \includegraphics[width=\columnwidth]{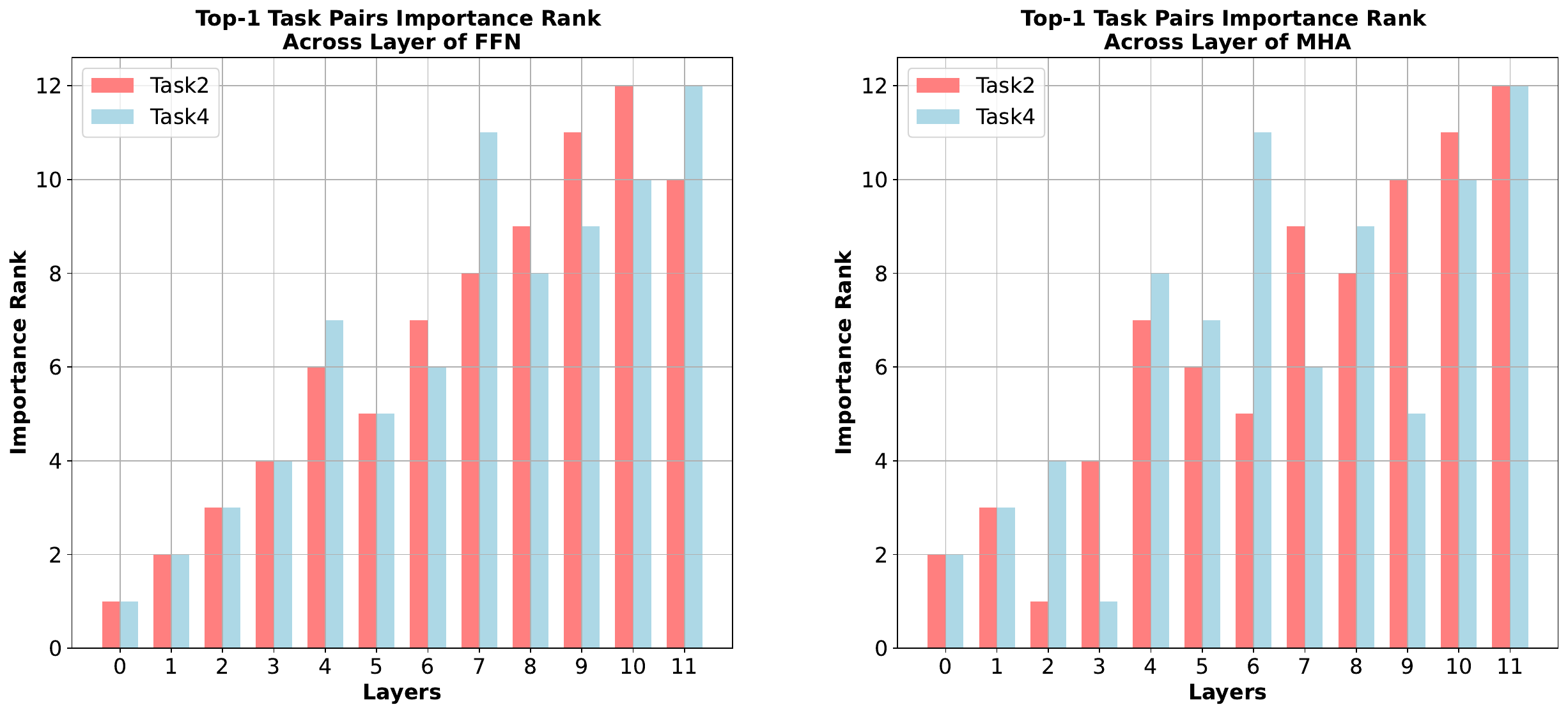}
  \caption{Comparison of layer-importance rankings across tasks.}
  \label{fig:Layer}
\end{figure}

Layer-Level Task Sensitivity.
We extend this analysis to the layer level. From the previous study, we identify the task pair exhibiting the greatest divergence in neuron-importance rankings. For this task pair, we compute KL-divergence–based layer-importance scores $\delta_l$ following the procedure described in Section~\ref{sec:Methodology}. These scores capture the sensitivity of each layer to pruning-induced perturbations in the model’s output distribution.
Figure~\ref{fig:Layer} compares the resulting layer-importance profiles of the two tasks. The layer rankings differ considerably, revealing substantial task-dependent variability in how ViT layers contribute to downstream predictions. This finding supports our central hypothesis: enforcing uniform or static layer budgets fails to account for heterogeneous layer sensitivities across tasks.

Together, the neuron- and layer-level analyses provide empirical evidence that both fine-grained (neuron) and coarse-grained (layer) structural importance in ViTs varies notably across tasks. This variability underscores the necessity of task-adaptive pruning. By dynamically adjusting pruning ratios according to task-specific sensitivities inferred from a lightweight metric dataset—without accessing device-local data—TAP-ViTs preserves essential capacity where it matters most for each deployment context.

\subsubsection{Component-wise Ablation Analysis} 

To thoroughly assess the contribution of each module within the TAP-ViTs framework, we perform a set of controlled ablation studies that dissect the roles of our two core components: (1) the GMM-based metric dataset construction, which provides a task-aware surrogate for local data, and (2) the efficient pruning strategy based on composite neuron importance and adaptive layer-wise sensitivity. Ablation experiments are conducted on both TinyImageNet and CIFAR-100 using DeiT-B and DeiT-S as representative backbones, and the results are summarized in Table~\ref{tab:ablation-components}.

The ablation results reveal several clear and consistent trends. First, removing the metric dataset construction leads to a dramatic drop in performance across all model–dataset combinations, underscoring its central role in enabling task-adaptive pruning without direct access to private data. For instance, on DeiT-B, accuracy decreases from 88.24\% to 81.12\% on TinyImageNet and from 92.72\% to 81.98\% on CIFAR-100. A similar decline is observed for DeiT-S, where accuracy drops by 8.98\% and 13.93\% on the two datasets, respectively. Such substantial degradation highlights that the constructed metric dataset effectively preserves class-discriminative structure and provides a robust, privacy-preserving proxy for guiding pruning decisions.

Second, removing the efficient pruning module—i.e., disabling composite neuron importance estimation and adaptive layer ranking—also results in noticeable performance degradation, though to a lesser extent than removing the dataset construction. For example, on DeiT-S, accuracy on TinyImageNet drops from 81.50\% to 79.49\%, and on CIFAR-100 from 90.32\% to 89.40\%. This confirms that our pruning strategy successfully identifies and retains task-critical components while removing redundancy more effectively than uniform or heuristic pruning methods.

% \begin{table}[t]
% \centering
% \caption{Detection performance of TAP-ViTs with PVT-Small backbone on COCO.}
% \renewcommand{\arraystretch}{1.3}
% \begin{tabular}{lcccccc}
% \toprule
% \textbf{Model} & \textbf{mAP} & \textbf{mAP$_{50}$} & \textbf{mAP$_{75}$} & \textbf{mAP$_s$} & \textbf{mAP$_m$} & \textbf{mAP$_l$} \\
% \midrule
% Original & 38.7 & 59.3 & 40.8 & 21.2 & 41.6 & 54.4 \\
% TAP-ViTs & 35.9 & 55.8 & 38.1 & 20.2 & 38.6 & 49.1 \\
% \bottomrule
% \end{tabular}

% \label{tab:detection}
% \end{table}

% \begin{figure}[t]
%     \centering
%     \hspace{-10mm}
%     \includegraphics[width=\linewidth]{time.pdf}
%     \caption{GMM Fitting Time on Device}
%     \label{fig:deployment-gmm}
% \end{figure}

Taken together, these observations demonstrate that the two components of TAP-ViTs are both essential and complementary. The GMM-based metric dataset construction provides the necessary task-awareness under data-constrained or privacy-sensitive settings, while the efficient pruning strategy delivers fine-grained adaptiveness that preserves performance even under substantial parameter reduction. Their combination yields a robust and highly effective pruning framework capable of generalizing across architectures and datasets.

% \subsubsection{Object Detection Results}
% To further evaluate the task generality of TAP-ViTs beyond classification, we apply our pruning strategy to a PVT-Small backbone and integrate the pruned model into a standard RetinaNet detector on the COCO dataset. We retain 80\% of the backbone parameters and fine-tune the detector for only 12 epochs following the common training protocol. The resulting performance is reported in Table~\ref{tab:detection}.

% Despite pruning 20\% of the parameters and performing only a short fine-tuning schedule, the pruned model maintains strong detection accuracy across all metrics. The relatively small gap in mAP, especially at the \emph{mAP$_s$} and \emph{mAP$_m$} scales, demonstrates that TAP-ViTs preserves the essential representational capacity required for dense prediction tasks. This result provides further evidence that our pruning framework is inherently task-agnostic and transfers effectively beyond classification.

\subsubsection{Real-World Deployment Analysis} 
\begin{figure}[t]
    \centering
    % \hspace{-6mm}
    \includegraphics[width=\linewidth, height=0.9\linewidth]{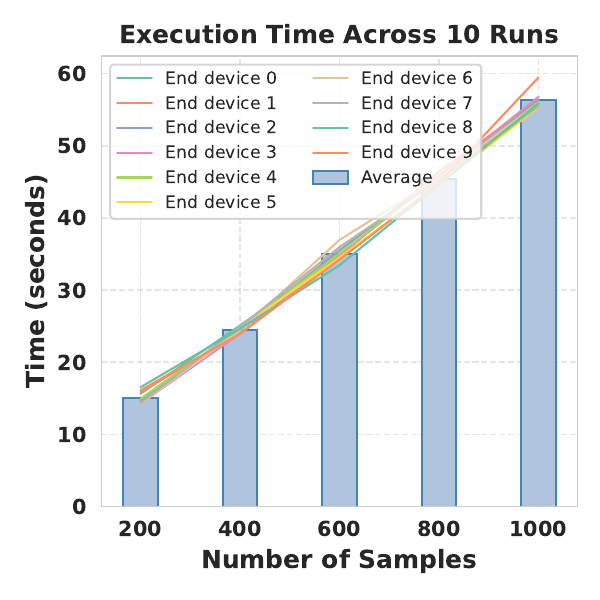}
    \caption{GMM Fitting Time on Device}
    \label{fig:deployment-gmm}
\end{figure}

% To evaluate the practicality of our approach in the real-world on-device deployment scenario, we evaluate TAP-ViTs on a real-world testbed. Specifically, we measure the time required to fit the local data distribution via GMM construction as this is the only step in the entire pruning process that needs to be executed on-device. As illustrated in Figure~\ref{fig:deployment-gmm}, the fitting time grows linearly with the amount of local data, rather than quadratically or exponentially. This linear trend indicates excellent scalability: even when the local dataset size increases substantially, the computational burden on the end device remains minimal. Across all tested settings, the fitting process completes in under one minute, highlighting the negligible cost introduced at the edge and underscoring the method’s suitability for real-world deployment scenarios. 
To assess the efficiency and feasibility of TAP-ViTs in realistic on-device deployment environments, we benchmark the end-to-end cost of executing the only device-side computation required by our framework: fitting a Gaussian Mixture Model (GMM) to approximate the local data distribution. This experiment is conducted on a real testbed, and the fitting time is measured over multiple runs for varying local sample sizes. The results are shown in Figure~\ref{fig:deployment-gmm}.

The figure reveals a clear and consistent linear relationship between the number of local samples and the GMM fitting time. Both the per-device curves and the averaged execution time increase proportionally with dataset size, demonstrating that the procedure scales gracefully without introducing quadratic or super-linear overhead. This behavior is particularly important for edge scenarios, where computational capacity is limited and latency constraints are stringent.
Moreover, the absolute execution time remains highly manageable. Even at the largest tested sample size (1000 samples), all devices complete the fitting process within approximately one minute. The tight clustering of the per-device curves also indicates low performance variability across devices, suggesting that the proposed metric dataset construction is robust to hardware heterogeneity.

Overall, these results validate the practicality of TAP-ViTs in real-world deployments. The lightweight nature of the on-device GMM fitting ensures that the entire pruning pipeline imposes negligible computational overhead at the edge, even on low-resource devices. This efficiency, combined with the framework’s ability to capture device-specific task characteristics without accessing any private data, makes TAP-ViTs a highly deployable solution for privacy-preserving and device-adaptive Vision Transformer compression. Such properties position TAP-ViTs as a practical component within broader edge intelligence systems, enabling scalable, secure, and high-quality model customization across heterogeneous deployment environments.

% Overall, these results validate the practicality of TAP-ViTs in real-world deployments. The lightweight nature of the on-device GMM fitting ensures that the entire pruning pipeline imposes negligible computational load at the edge, making TAP-ViTs a highly deployable solution for privacy-preserving and device-adaptive vision transformer compression.

\section{Conclusion and Discussion} \label{sec:Conclusion and Discussion}
% We proposed TAP-ViTs, a task-adaptive pruning framework for Vision Transformers that enables customized model compression without accessing local data. By constructing a GMM-based metric dataset from public data and introducing a dual-level importance evaluation strategy, TAP-ViTs effectively captures task-specific requirements while preserving data privacy. Extensive experiments across diverse backbones and datasets demonstrate its superiority over state-of-the-art methods, achieving minimal accuracy loss even under aggressive pruning. The framework’s task adaptivity and real-world deployability make it a promising solution for privacy-aware on-device AI. We plan to extend TAP-ViTs to multi-modal domains and integrating it with complementary compression techniques in future works.

% While TAP-ViTs preserves data privacy by avoiding access to raw local data, it relies on the availability of large-scale, high-quality public datasets on the cloud to approximate private task distributions. Future work could explore generating more accurate proxy datasets on the cloud using generative foundation models. In terms of social impact, the framework enhances equitable access to efficient AI on low-resource end devices, potentially reducing digital inequality. However, as with all model compression techniques, care must be taken to ensure that task-specific pruning does not unintentionally amplify biases or degrade performance on underrepresented groups. 

This work introduced TAP-ViTs, a task-adaptive pruning framework for Vision Transformers that enables device-specific model compression without accessing any raw local data. By constructing a lightweight GMM-based approximation of each device’s data distribution and sampling a matched metric dataset from public cloud data, TAP-ViTs provides a principled mechanism for capturing device-specific task characteristics while preserving data privacy. Building on this foundation, we further developed a dual-level importance evaluation strategy—comprising composite neuron importance estimation and adaptive layer importance assessment—to guide fine-grained, task-aware pruning decisions with high efficiency.

Extensive experiments across multiple ViT architectures and datasets consistently demonstrate the effectiveness and robustness of TAP-ViTs. The framework achieves minimal accuracy degradation even under substantial parameter reduction, and under low pruning ratios, the customized models can even surpass the performance of unpruned baselines, reflecting their reduced redundancy and enhanced task alignment. These results highlight the practicality of TAP-ViTs as a drop-in solution for resource-constrained and privacy-sensitive on-device AI deployment.

Looking forward, several extensions merit exploration. First, although TAP-ViTs avoids direct exposure of private data, it currently assumes the availability of sufficiently rich public datasets for constructing proxy metric datasets. Future work hopes to explore leveraging generative foundation models to synthesize more accurate and diverse proxy samples directly in the cloud, further narrowing the gap between public and private data distributions. Second, generalizing the framework to multi-modal settings (e.g., vision–language models or video–audio models) may extend its applicability to a broader range of edge intelligence scenarios.

From a societal perspective, the proposed framework has the potential to enhance equitable access to efficient AI capabilities on low-resource edge devices, particularly in regions where computational infrastructures remain limited. By enabling privacy-preserving yet high-quality model customization, TAP-ViTs may help reduce digital inequality and promote more inclusive AI technologies. At the same time, as with all pruning and model compression approaches, care must be taken to ensure that task-adaptive pruning does not inadvertently amplify model biases or degrade performance on underrepresented demographic groups. Ongoing evaluation and responsible deployment practices will be essential to ensure that such techniques contribute positively to broader societal goals and to fostering cooperation across global AI ecosystems.

\bibliographystyle{unsrt}
\bibliography{references}

\end{document}